\documentclass{article}

\usepackage{microtype}
\usepackage{graphicx}
\usepackage{subfigure}
\usepackage{booktabs} 

\usepackage{hyperref}

\usepackage[arxiv]{icml2024}

\usepackage{amsmath}
\usepackage{amssymb}
\usepackage{mathtools}
\usepackage{amsthm}

\usepackage[capitalize,noabbrev]{cleveref}

\usepackage{bm}
\usepackage{textcomp}
\usepackage{tablefootnote}
\usepackage{multirow}
\usepackage{subcaption}
\usepackage{makecell}

\newcommand{\inputspace}{\mathcal{X}}
\newcommand{\latentspace}{\mathcal{Z}}

\newcommand{\x}{\bm{\mathrm{x}}}
\newcommand{\y}{\mathrm{y}}
\newcommand{\z}{\bm{\mathrm{z}}}
\newcommand{\xopt}{\x^*}

\newcommand{\X}{\bm{\mathrm{X}}}
\newcommand{\Y}{\bm{\y}}
\newcommand{\Z}{\bm{\mathrm{Z}}}

\newcommand{\kernelfunc}{\kappa_0}

\newcommand{\zunlabel}{\z_u}

\newcommand{\xsetlabel}{\X_l}
\newcommand{\zsetlabel}{\Z_l}
\newcommand{\ysetlabel}{\Y_l}

\newcommand{\zsetunlabel}{\Z_u}
\newcommand{\ysetunlabel}{\hat{\Y}_u}

\newcommand{\pseudolabels}{\hat{\Y}_u}
\newcommand{\Teacher}{T}
\newcommand{\Student}{S}

\newcommand{\lrU}{\eta_{u}}
\newcommand{\acq}{\alpha}
\newcommand{\rv}{\bm{\mathrm{r}}}
\newcommand{\rvset}{\bm{\mathrm{R}}}
\newcommand{\Encoder}{\psi}
\newcommand{\Decoder}{\varphi}
\newcommand{\datasetlabel}{\mathcal{D}_{l}}
\newcommand{\datasetval}{\mathcal{D}_{v}}
\newcommand{\datasetunlabel}{\mathcal{D}_{u}}

\newcommand{\R}{\mathbb{R}}
\newcommand{\pdfZu}{p_{\z_u}}

\newcommand{\pdfrv}{p_{\rv}}
\newcommand{\topk}{K}

\newcommand{\ourmethod}{\texttt{TSBO}}

\newcommand{\paramsT}{\bm{\theta_{\Teacher}}}
\newcommand{\paramsS}{\bm{\theta_{\Student}}}

\newcommand{\paramsU}{\bm{\theta_{u}}}

\newcommand{\gpmodel}{Q}

\newcommand{\Mean}{\bm{\mu}}
\newcommand{\Var}{\bm{\Sigma}}
\newcommand{\teacherPredMean}{\Mean_{\paramsT}}
\newcommand{\teacherPredVar}{\Var_{\paramsT}}
\newcommand{\studentPredMean}{\Mean_{\paramsS}}

\newcommand{\E}{\mathbb{E}}

\newcommand{\losslabeled}{\mathcal{L}_{l}}
\newcommand{\lossunlabeled}{\mathcal{L}_{u}}
\newcommand{\lossfeedback}{\mathcal{L}_{f}}

\DeclareMathOperator*{\argmax}{argmax}
\DeclareMathOperator*{\argmin}{argmin}

\makeatletter
\newcommand{\smallbullet}{} 
\DeclareRobustCommand\smallbullet{%
  \mathord{\mathpalette\smallbullet@{0.7}}%
}
\newcommand{\smallbullet@}[2]{%
  \vcenter{\hbox{\scalebox{#2}{$\m@th#1\bullet$}}}%
}
\makeatother

\theoremstyle{plain}

\theoremstyle{definition}

\theoremstyle{remark}

\usepackage[textsize=tiny]{todonotes}

\icmltitlerunning{High-Dimensional Bayesian Optimization via Semi-Supervised Learning with Optimized  Unlabeled Data Sampling}

\begin{document}

\twocolumn[
\icmltitle{High-Dimensional Bayesian Optimization via Semi-Supervised Learning \\ with Optimized  Unlabeled Data Sampling}



\icmlsetsymbol{equal}{*}

\begin{icmlauthorlist}
\icmlauthor{Yuxuan Yin}{yyy}
\icmlauthor{Yu Wang}{yyy}
\icmlauthor{Peng Li}{yyy}
\end{icmlauthorlist}

\icmlaffiliation{yyy}{Department of ECE, University of California, Santa Barbara, USA}

\icmlcorrespondingauthor{Yuxuan Yin}{y\_yin@ucsb.edu}

\icmlkeywords{Machine Learning, ICML}

\vskip 0.3in
]



\printAffiliationsAndNotice{}  

\begin{abstract}
We introduce a novel semi-supervised learning approach, named Teacher-Student Bayesian Optimization ($\texttt{TSBO}$), integrating the teacher-student paradigm into BO to minimize expensive labeled data queries for the first time.
$\texttt{TSBO}$ incorporates a teacher model, an unlabeled data sampler, and a student model. The student is trained on unlabeled data locations generated by the sampler, with pseudo labels predicted by the teacher. The interplay between these three components implements a unique \emph{selective regularization} to the teacher in the form of student feedback. This scheme enables the teacher to predict high-quality pseudo labels, enhancing the generalization of the GP surrogate model in the search space. To fully exploit $\texttt{TSBO}$, we propose two optimized unlabeled data samplers to construct effective student feedback that well aligns with the objective of Bayesian optimization. Furthermore,  we quantify and leverage the uncertainty of the teacher-student model for the provision of reliable feedback to the teacher in the presence of risky pseudo-label predictions. $\texttt{TSBO}$ demonstrates significantly improved sample-efficiency in several global optimization tasks under tight labeled data budgets. 


\end{abstract}

\section{Introduction}\label{sec:Introduction}
Bayesian Optimization (BO) \citep{brochu2010bo} is widely adopted for black-box optimization, which is particularly useful when the objective function is expensive or impractical to evaluate directly. BO operates by constructing a surrogate model, e.g., a Gaussian Process (GP) \citep{seeger2004gp}, of the objective function and then iteratively selecting the most promising locations for new labeled data query based on a criterion that balances exploration and exploitation \cite{kushner1964pi, jones1998ei, srinivas2009ucb}. 
Recent work has extended BO's applicability to high-dimensional tasks with various dimension reduction methods including linear embedding \citep{wang2016rembo, chen2020silbo}, nonlinear projections \citep{moriconi2020high-bo}, and deep autoencoders  \citep{kusner2017grammar-vae,jin2018jtvae, tripp2020, chen2020silbo, grosnit2021tlbo, lol-bo, pg-lbo}.

Despite these encouraging advances, labeled data acquisition remains inherently costly and presents a key bottleneck in BO across many application domains such as functional molecule design \cite{brown2019guacamol, gao2022sample-efficiency-matters}, structural optimization \citep{zoph2018nas, lukasik2022nas}, and failure analysis \citep{hu2018failure-detection, liang2019failue-detection}.


\begin{figure}[tbp]
    \centering
    \includegraphics[width=\columnwidth,trim=2 4 2 2,clip]{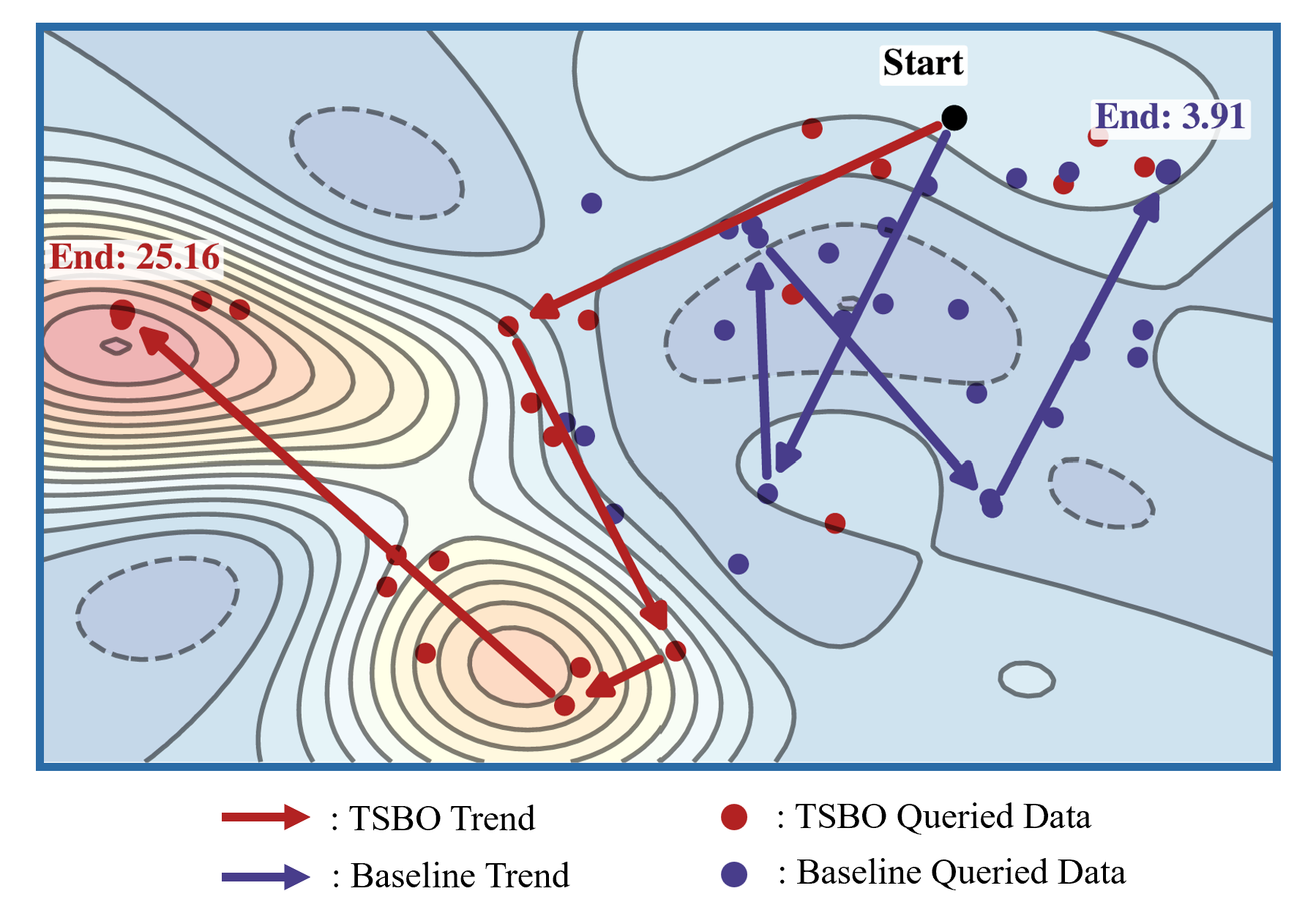}
    \caption{Visualization of queried data (dots) and trends (arrow sequences) on a high-dimensional molecule design task \citep{sterling2015zinc} to maximize the Penalized LogP score \citep{gomez2018semibo1}. Red and blue colors represent \ourmethod\ and a baseline (with vanilla BO), respectively. The evaluation budget is 450 in both approaches.}
    \label{fig:visual-umap}
\end{figure}

\begin{figure*}[htbp]
    \centering
    \includegraphics[width=1.0\textwidth,trim=2 4 2 2,clip]{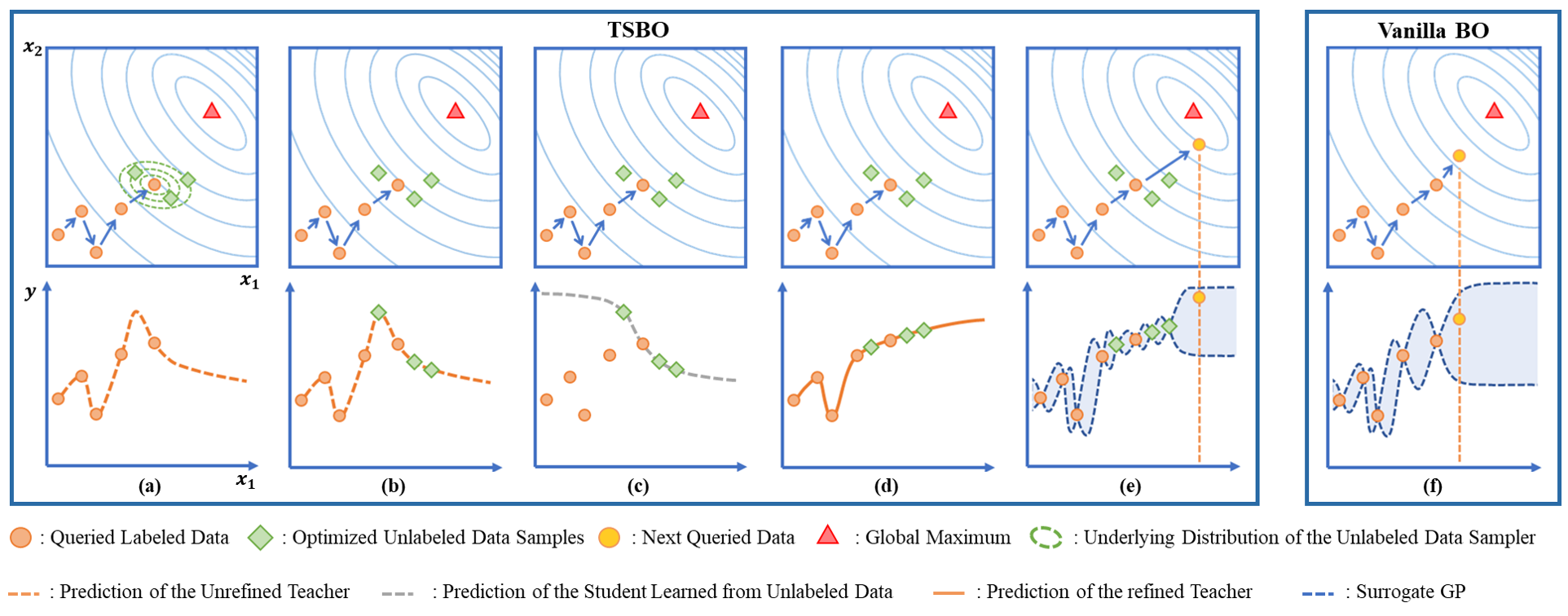}
    \caption{Illustrated example to demonstrate the interaction between the unlabeled data sampler, the teacher and the student employs selective regularization. (a): unlabeled data are sampled from regions with potentially high values. (b): the teacher predicts pseudo labels for unlabeled data. (c): the student learns from the unlabeled data and the predicted pseudo labels and is evaluated on labeled data as the feedback. (d): the teacher refines its prediction based on the feedback. (e): GP in $\ourmethod$ fits on both the labeled data and unlabeled data with refined pseudo labels. (f): GP in vanilla BO fits only on the labeled data.}
    \label{fig:tsbo_flow}
\end{figure*}

To address the challenge of data query efficiency in black-box optimization, we introduce a unified Semi-Supervised Learning (SSL) approach called Teacher-Student Bayesian Optimization ($\ourmethod$). $\ourmethod$ is the first work integrating a teacher-student model into BO, and bridges the gap between SSL and the goals of BO by implementing a unique \emph{selective regularization} mechanism, allowing the use of a subset of ``potentially high-quality" unlabeled data from a vast sample space. This targeted utilization of cheap unlabeled data is optimized to serve the BO's optimization objective, steering it more effectively towards areas of high values with dramatically increased data query efficiency compared with baseline methods. An illustrated example on $\ourmethod$'s sequential optimization process is shown in \cref{fig:visual-umap}, 
which demonstrates the superior sample efficiency of \ourmethod. 

At its core, $\ourmethod$ incorporates a \emph{teacher model}, a \emph{student model}, and an \emph{unlabeled data sampler}. It is the interplay of the three components that implement our targeted  \emph{selective regularization} to the teacher. Ultimately, the regularized teacher predicts high-quality pseudo labels \footnote{ Although the term ``label" often means the ground truth in classification problems, it is also widely used to represent observed values in BO tasks \citep{grosnit2021tlbo,chen2020silbo, jean2018semi-gp}.}, which supplement the queried labeled data to better train the standard GP surrogate model for more optimized new labeled data query during Bayesian optimization.  

As illustrated in \cref{fig:tsbo_flow}, at each BO iteration,  the unlabeled data sampler selects a set of \emph{optimized} unlabeled data locations and passes them onto the teacher, which utilizes its current knowledge to prediction pseudo labels \citep{lee2013pseudo-labeling, pham2021meta_pseudo_label} for them. The student is then trained exclusively on the pseudo labels predicted by the teacher. Recognizing the fact the teacher's pseudo-label prediction can be misleading, the student is evaluated on the existing ground truth labeled data, and its performance is fed back to the teacher. Subsequently, the teacher is refined with the student feedback included as the selective regularization. Upon the completion of teacher-student interaction, the teacher refined with the regularization provides reliable pseudo-labels for training the BP surrogate model, bolstering its new data query capability even with limited labeled training data.  




To fully exploit $\ourmethod$, it is essential to carefully design the key components involved in selective regularization. Instead of employing random sampling of unlabeled data, our unlabeled data sampler strategically places unlabeled data in regions of high-quality pseudo labels, and simultaneously encourages exploration towards the global optimum. Providing the student feedback to the teacher based on unlabeled data sampled under this strategy evaluates and refines the teacher in a way that well aligns with the overall objective of Bayesian optimization. We propose two optimized unlabeled data samplers: one based on the Extreme Value Theory (EVT) \citep{fisher1928evt} and the other on a parameterized sampling distribution. 

\begin{figure*}[tbp]
\centering
\includegraphics[width=0.95\textwidth,trim=2 10 2 2,clip]{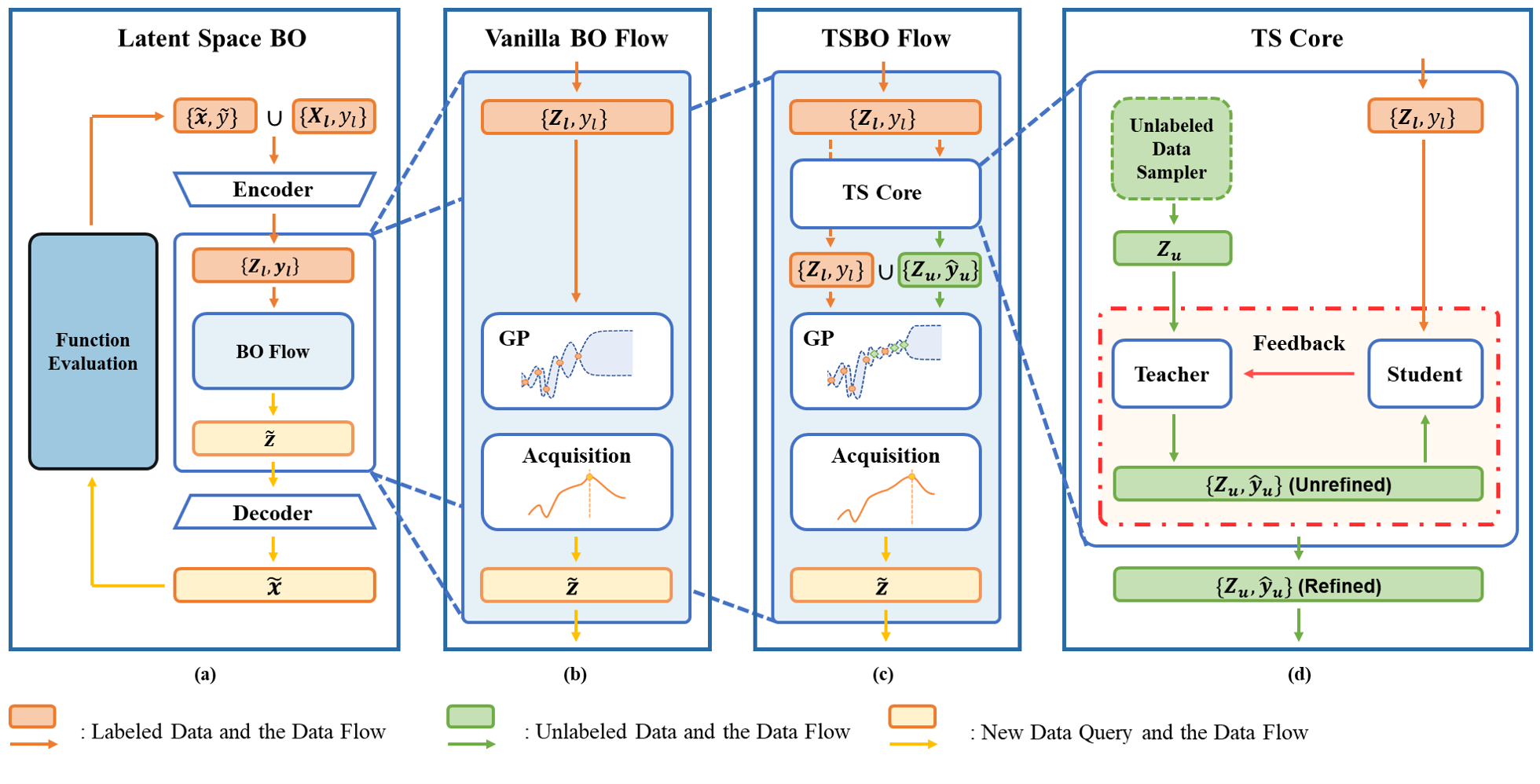} 
\caption{Overview of the $\ourmethod$ framework. (a): the basic Latent Space BO architecture. (b): the vanilla BO flow utilizes only the encoded labeled data to train the GP model and query the next data. (c): $\ourmethod$ flow incorporates a TS Core to provide additional high-quality unlabeled data to the GP model during each BO iteration.. (d): inside TS Core: the optimized unlabeled data sampler and the feedback from the student provides selective regularization to the teacher.}
\label{fig:tsbo}
\end{figure*}

Furthermore, we boost the performance of $\ourmethod$ by making our teacher-student model \emph{uncertainty aware}. The teacher not only assigns pseudo labels to the selected unlabeled data but also quantifies their uncertainty. These uncertainties are taken into account during the student training process,  thereby providing more reliable feedback to the teacher, especially in the presence of the risk associated with unreliable pseudo-label predictions.



We evaluate the proposed $\ourmethod$ on various challenging high dimensional datasets and show superior data efficiency improvement. In a chemical design task \citep{sterling2015zinc} and an expression reconstruction task \citep{kusner2017grammar-vae}, we respectively achieve SOTA results within 3\textperthousand\ and 1\% evaluations compared to recent works.

\section{Preliminaries}

\subsection{Bayesian Optimization (BO)}



Bayesian Optimization (BO) \citep{brochu2010bo} is designed to identify the global maximum of an unknown, typically non-convex function $f: \inputspace \to \R$, defined as:
\begin{equation} \label{eq:obj}
\xopt = \argmax_{\x \in \inputspace} f(\x)
\end{equation}
where \(\inputspace \subseteq \R^D \) represents a $D$-dimensional input space.

BO approaches this challenge by iteratively selecting data points $\x$ for evaluation, building upon the outcomes of previously queried points. Within a single BO iteration, given a set of $N$ evaluated examples $\{\x_i,\y_i\}_{i=1}^N = \{\xsetlabel,\ysetlabel\}$, where $\xsetlabel$ is an $N\times D$ matrix of inputs and $\ysetlabel$ is an $N \times 1$ vector of corresponding outputs, the next query point $\Tilde{\x}$ is determined by:
\begin{equation} \label{eq: bo step}
    \Tilde{\x} = \argmax_{\x\in \inputspace} \acq \Big( \gpmodel \big(\x | \{ \xsetlabel,\ysetlabel \} \big) \Big)
\end{equation}
Here, $\gpmodel \big(\x | { \xsetlabel,\ysetlabel } \big)$ denotes the posterior distribution of $\x$ estimated by a probabilistic model $\gpmodel$, commonly termed as surrogate model\citep{frazier2018bo-tutorial}. The acquisition function $\acq$ is designed to strike a balance between exploiting regions of known high values and exploring unknown spaces.


\subsection{Latent Space BO} 
For high-dimensional challenges, direct application of Bayesian Optimization (BO) in the original space can lead to overfitting in the data query model $\gpmodel$, particularly under a constrained data query budget, a dilemma often referred to as the curse of dimensionality \citep{brochu2010bo}. A viable remedy is latent space BO \citep{gomez2018semibo1, kusner2017grammar-vae}, which conducts BO within a generative latent space $\latentspace \subseteq \R^d$ where $d \ll D$. This approach utilizes an encoder $\Encoder: \inputspace \to \latentspace$ and a decoder $\Decoder:\latentspace \to \inputspace$ to navigate the query process. In latent space BO, the next query ${ \Tilde{\x}, \Tilde{\y}}$, informed by the current labeled data ${\xsetlabel,\ysetlabel}$, involves:

$\smallbullet\ $ Fitting the data query model $\gpmodel$ in $\latentspace$ using latent code pairs ${\zsetlabel,\ysetlabel}$, where $\zsetlabel:= \Encoder(\xsetlabel)$;

$\smallbullet\ $ Identifying an optimal latent code $\Tilde{\z}$ by maximizing the acquisition function $\acq$ in $\latentspace$: $\Tilde{\z}= \argmax_{\latentspace} \acq \Big( \gpmodel \big(\z | { \zsetlabel,\ysetlabel } \big) \Big)$;

$\smallbullet\ $ Decoding $\Tilde{\x} = \Decoder(\Tilde{\z})$ and evaluating $\Tilde{\y} = f(\Tilde{\x})$.

The basic latent space BO framework is visualized in \cref{fig:tsbo} (a).

\section{$\ourmethod$ Problem Formulation}\label{sec:problem-formulation}
The $\ourmethod$ framework builds upon the conventional latent space BO to enhance data efficiency in challenging high-dimensional tasks. While retaining the encoder-decoder structure and following the established data query procedure, $\ourmethod$ additionally introduces a teacher model, a student model, and an unlabeled data sampler to employ selective regularization in the latent space between each data query. This key step enriches the surrogate model with a wealth of high-quality unlabeled data and their corresponding calibrated pseudo labels, thereby enhancing the model's predictive performance, as depicted in \cref{fig:tsbo} (c) and \cref{fig:tsbo} (d).

Let $T(\cdot;\bm{\theta_{T}})$ represent the teacher model, $S(\cdot;\bm{\theta_{S}})$ the student model, and $\pdfZu(\cdot;\bm{\theta_{u}})$ the distribution underlying the unlabeled data sampler.
The set $\datasetlabel:=
\{\zsetlabel,\ysetlabel\}$ denotes the labeled data available at the current step. The interaction between these modules is formally described below and formulated as a bi-level optimization problem:

\begin{itemize}
    \item [1.] Unlabeled data $\zsetunlabel$ is sampled from $\pdfZu(\cdot;\bm{\theta_{u}})$.
    \item [2.] The teacher predicts pseudo labels on the unlabeled data samples: $\pseudolabels (\paramsT) := T(\zsetunlabel;\bm{\theta_{T}})$.
    \item [3.] The student is trained with an unlabeled loss $\mathcal{L}_{u}$ defined on an unlabeled dataset $\datasetunlabel(\paramsT): =\{\zsetunlabel, \ysetunlabel\}$:
    \begin{equation} \label{eq:def-student-obj}
        \bm{\theta_{S}^{*}}(\bm{\theta_{T}}) = \argmin_{\paramsS}  \lossunlabeled\big( \datasetunlabel(\paramsT); \paramsS \big)
    \end{equation}
     \item [4.] The trained student is evaluated on the labeled dataset $\datasetlabel$ by a feedback loss $\lossfeedback\big (\datasetlabel; \paramsS^*(\paramsT) \big)$
   \item [5.] The teacher is updated by optimizing the combination of a labeled loss $\mathcal{L}_{l}$ and the feedback loss $\lossfeedback$:
    \begin{equation} \label{eq:def-teacher-obj}
        \paramsT^* = \argmin_{\paramsT}   \losslabeled(\datasetlabel;\paramsT)  + 
         \lambda \lossfeedback \big (\datasetlabel; \paramsS^*(\paramsT) \big)
    \end{equation}
    where $\lambda$ is a weighting parameter.
\end{itemize}


\section{Detailed Design of TSBO Modules}
\subsection{Uncertainty-Aware Teacher-Student Model}\label{sec:un-aw-tsmodel}

\begin{figure*}[t]
\centering
\includegraphics[width=1.0\textwidth]{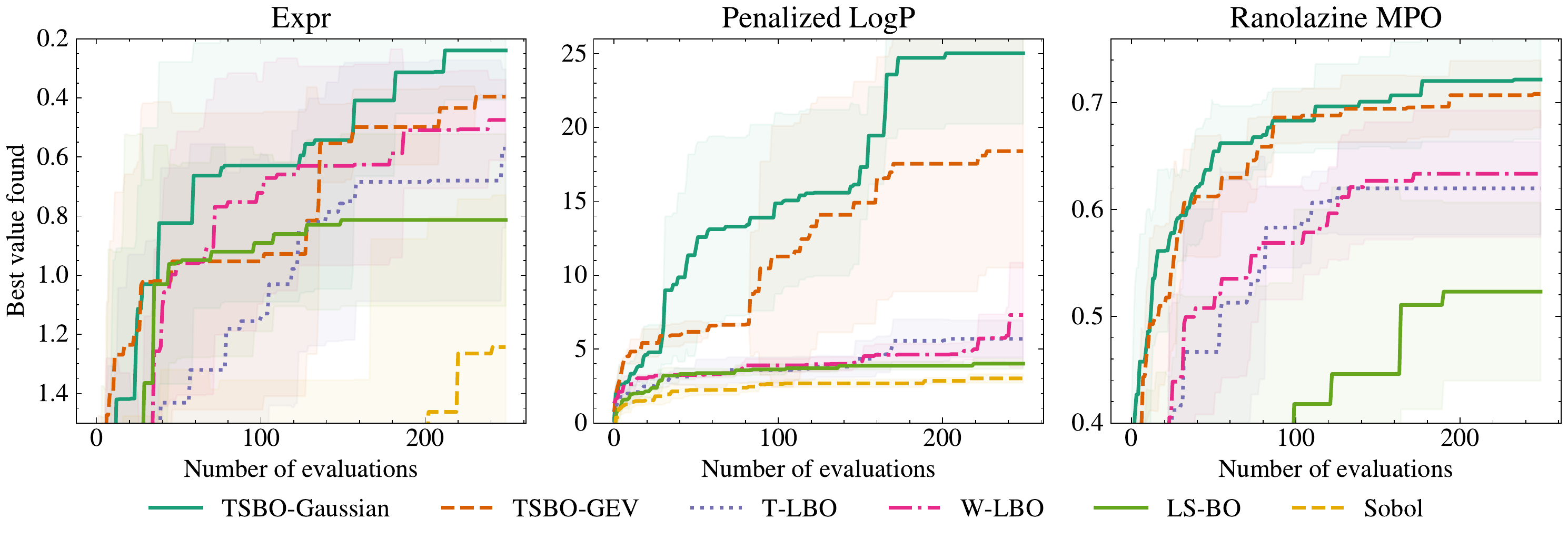} 
\caption{Comparison between the mean performance and standard deviations between 4 LSO baselines and \ourmethod. }
\label{fig:main_results}
\end{figure*}

The interplay between the teacher and student models is pivotal for refining the teacher's predictions and enhancing the accuracy of pseudo labels on the sampled unlabeled data. Crucially, we emphasize the importance of incorporating uncertainty awareness to safeguard the feedback mechanism's integrity. Inaccurate pseudo labels can disrupt the modeling process, leading to erroneous feedback. By quantifying prediction uncertainty, the student model can more prudently utilize pseudo labels, adjusting its reliance on them based on their associated uncertainty levels. As a result, we propose the uncertainty-aware teacher-student, a practical mechanism to utilize pseudo labels effectively while minimizing potential risks.

\subsubsection{Uncertainty-Aware Teacher Model}
Quantifying uncertainty accurately is inherently complex and computationally demanding. In $\ourmethod$, we adopt a heuristic approach inspired by \cite{mlp-teacher} using a Multilayer Perceptron (MLP) as the teacher model. This MLP is designed to output not only the predicted mean but also the diagonal covariance matrix for any given input $\Z$, thereby providing a probabilistic estimate of the output:
\begin{align}
    \Teacher(\Z;\paramsT) := \mathcal{N} (\teacherPredMean (\Z), \teacherPredVar (\Z)), \\
     \text{where}\quad \teacherPredVar (\Z):=\bm{\sigma_{\paramsT}^{2}}(\Z)\cdot \bm{I}
\end{align} 
Training of the defined teacher model becomes fitting a parameterized Gaussian distribution by minimizing a negative log-likelihood (NLL) loss. Consequently, the labeled loss in \cref{eq:def-teacher-obj} is written as:
\begin{equation}
    \losslabeled(\datasetlabel;\paramsT) := \text{NLL} \big( \Teacher (\zsetlabel;\paramsT), \ysetlabel \big)
\end{equation}

The pseudo label that incorporated with uncertainty from the teacher is generated as follows:
\begin{equation}
\begin{aligned}
    &\hat{\y}_u = \y_u + \epsilon_u (\zunlabel), \\
   \text{where}\quad &\epsilon_u (\zunlabel) \sim \mathcal{N}(\bm{0},\teacherPredVar(\zunlabel))
\end{aligned}
\end{equation}

\subsubsection{Uncertainty-Aware Student Model}
In our approach, the student model is implemented as a Gaussian Process (GP) as in \cref{eq:student-prior}, which can efficiently incorporate the teacher's uncertainty predictions as prior knowledge. This configuration enables the student model not only to refine its predictions but also to provide precise feedback on the teacher model's mean predictions and associated uncertainties.
\begin{equation} \label{eq:student-prior}
    S(\cdot;\paramsS) := \mathcal{GP}(\bm{0}, \kappa_0(\cdot,\cdot) + \sigma_0^2 \bm{I})
\end{equation}
where $ \kappa_0(\cdot,\cdot))$ is a Radial Basis Function (RBF) kernel, $\sigma_0^2$ is an additive variance. The student model parameter $\paramsS$ encloses the kernel parameters and the variance $\sigma_0^2$.

We propagate the teacher's uncertainty $\epsilon_u (\zunlabel)$ to the downstream training of the  student GP model by forming a student's pseudo-label dependent prior: $\hat{\y}_u =  \epsilon_u (\zunlabel)  + \epsilon_{\kappa0}$, where
$\epsilon_{\kappa0} \sim \mathcal{N}(0,\kappa_0(\zunlabel,\zunlabel) + \sigma_0^2)$ is the student's vanilla prior variance in \cref{eq:student-prior}. The optimization of student's variance $\epsilon_{\kappa0}$ depends on the teacher's uncertainty estimation  $\teacherPredVar(\zunlabel)$. 

Correspondingly, the student's covariance matrix $\Var_u$ over the unlabeled dataset $\datasetunlabel(\paramsT)$ is computed as: $\Var_{u_{ij}} = \E( \hat{\y}_{u_i} - \E\ \hat{\y}_{u_i}) (\hat{\y}_{u_j} - \E\ \hat{\y}_{u_j}) = \kernelfunc(\z_{u_i}, \z_{u_j}) + \delta_{ij} \sigma_0^2 + \delta_{ij}\teacherPredVar(\z_{u_i})$ , where $\delta$ represents the Kronecker delta. It is the sum of the uncertainty of the teacher and the student:
\begin{equation}\label{eq:covariance} 
    \Var_u =  \kernelfunc(\zsetunlabel, \zsetunlabel) + \sigma_0^2 \bm{I} + \teacherPredVar(\zsetunlabel)
\end{equation}

We optimize the uncertainty-aware student's parameter $\paramsS$ on the unlabeled dataset $\datasetunlabel(\paramsT)$ to minimize the NLL loss $\lossunlabeled$:
\begin{equation}\label{eq:unlabled-loss}
    \lossunlabeled( \datasetunlabel(\paramsT) ; \paramsS) 
     := \text{NLL} \Big( \mathcal{N} \big(\bm{0}, \Var_u\big), \pseudolabels \Big)
\end{equation}
\subsubsection{Derivation of the Feedback Loss}

The student GP model which is optimized by minimizing the unlabeled data loss in \cref{eq:unlabled-loss}, is evaluated on the labeled dataset $D_{l}$. This evaluation yields a \textit{posterior} prediction for the encoded inputs $\Z_{l}$. We focus on the posterior mean $\studentPredMean(\zsetlabel ; \datasetunlabel(\paramsT))$ to quantify the model's performance and define the feedback loss as the Mean Squared Error (MSE)\footnote{The feedback loss can be an MSE, a negative predictive marginal log-likelihood \citep{gneiting2007npll-feedback} or a negative Mahalanobis distance \citep{bastos2009mahadist-feedback}. For numerical stability, we choose the MSE in our work.} between this posterior mean and the true labels $\ysetlabel$.
\begin{equation} \label{eq:feedbackloss}
        \lossfeedback \big (\datasetlabel; \paramsS \big) := \text{MSE} \Big ( \studentPredMean(\zsetlabel ; \datasetunlabel(\paramsT) ), \ysetlabel \Big)
\end{equation}
The posterior mean has an explicit form which is derived as:
\begin{equation} \label{eq:gp_post}
        \studentPredMean(\zsetlabel ; \datasetunlabel(\paramsT) ) = \kernelfunc(\zsetlabel, \zsetunlabel)^T {\Var_u}^{-1} \pseudolabels
\end{equation}

\paragraph{Remark:} The student GP model efficiently incorporates the teacher's uncertainty into its feedback, ensuring that its decisions are informed by this crucial context. Specifically, the model's posterior predictions show a diminished dependency on pseudo labels that are associated with high levels of uncertainty from the teacher: The teacher's predictive variance $\teacherPredVar(\z_{u_i})$ for the $i$-th pseudo label is added to the $i$-th diagonal entry of the covariance matrix $\Var_u$ in \cref{eq:covariance}. When this uncertainty is significantly greater than that of other pseudo labels, the corresponding diagonal element $\Var_{u_{ii}}$ is much larger than the other diagonal elements. As a result, per \cref{eq:gp_post} the contributions of the $i$-th pseudo label in $\pseudolabels$ to the posterior mean predictions of the validation labels are considerably reduced. 

\subsection{Optimized Unlabeled Data Sampling}\label{sec:unlabeled-data-sampling}
\begin{table*}[t]
    \centering
    \caption{Mean and standard deviation of the best value found after 250 data queries. }
    
    \begin{tabular}{l c c c}
    \toprule
    Method  &Expression ( $\downarrow$ ) & Penalized LogP ( $\uparrow$ ) & Ranolazine MPO ( $\uparrow$ )\\
    \midrule
    Sobol \citep{owen2003sobol}         &1.261±0.689 &3.019±0.296 & 0.260±0.046\\
    LS-BO \citep{gomez2018semibo1}      &0.579±0.356 &4.019±0.366 & 0.523±0.084\\
    W-LBO \citep{tripp2020}             &0.475±0.137 &7.306±3.551 & 0.633±0.059\\
    T-LBO \citep{grosnit2021tlbo}       &0.572±0.268 &5.695±1.254 & 0.620±0.043\\
    
    \midrule
    TSBO-GEV                            &0.396±0.070 &18.40±7.890 & 0.708±0.032\\
    TSBO-Gaussian                       &\textbf{0.240±0.168} &\textbf{25.02±4.794} & \textbf{0.744±0.030}\\
    
    \bottomrule
    \end{tabular}
    \label{tab:final}
\end{table*}

Random sampling, often used in standard Semi-Supervised Learning (SSL) scenarios, falls short in Bayesian Optimization (BO) due to its non-optimized nature. The teacher model's pseudo labels for randomly chosen unlabeled data, particularly those far  from the training dataset, are prone to low quality, characterized by small means or large variances. These pseudo labels can potentially mislead the student and steer it away from the global optimum. Consequently, an inaccurately informed student model will fail to provide the necessary feedback for the teacher model's refinement towards optimal solutions. 

We argue that within the vast expanse of the sample space, only a specific subset of unlabeled data holds significant value and aligns with the BO objectives. Utilizing these unlabeled data should benefit the BO process most effectively. In light of this, the development of a targeted unlabeled data sampling strategy, one that efficiently identifies and selects these high-value samples, is crucial. To meet this objective, we propose two novel techniques designed to enhance the sampling distribution for unlabeled data, ensuring a more focused and effective optimization in BO settings.


\paragraph{Method 1: Extreme Value Theory (EVT) based Unlabeled Data Sampling} \label{sec:gev}


The key idea is to place unlabeled data in regions of high-quality pseudo labels and at the same time encourage exploration towards the global optimum. To do so, we model the distribution of the part of the labeled data, which are extreme, i.e. with the best target values. EVT \citep{fisher1928evt} states that if $\{\y_1, \cdots, \y_N\}$ are i.i.d. and as $N$ approaches infinity, their maximum $\y^*$ follows a Generalized Extreme Value (GEV) distribution \citep{fisher1928evt}. We discuss the optimization and the Markov-Chain Monte-Carlo (MCMC) \citep{Andrieu:2003fk} sampling approach of GEV \citep{hu2019gevmcmc} in \cref{sec:appendix-gev-sampling}.

\paragraph{Method 2: Unlabeled Data Sampling Distribution Learned from Student's Feedback}\label{sec:unlabel-update-rule}
While the proposed GEV distribution approach offers a theoretically sound method for generating unlabeled data, its practical effectiveness is constrained by the computationally intensive nature of the MCMC sampling technique. 

To circumvent the computational burden associated with MCMC, we endeavor to identify an alternative approach for sampling unlabeled data, denoted as $\zunlabel$, from a distribution $\pdfZu (\cdot;\paramsU)$  parameterized $\paramsU$. We propose to optimize the $\pdfZu$ by minimizing the feedback loss $\lossfeedback$. 
Intuitively, a large $\lossfeedback$ is indicative of the use of unlabeled data with poor pseudo-label quality, which can potentially mislead the teacher-student model. In practice, We adopt the reparametrization trick \citep{kingma2013vae} to optimize $\paramsU$, and integrate it into the training loop of the teacher-student model. Details are elaborated in \cref{sec:appendix-unlabel-update-rule}.

\section{Experiments}\label{sec:results}
\begin{table*}[tb]
    \centering
    \caption{A broader comparison on the Chemical Design Task to maximize the Penalized LogP}
    
    \begin{tabular}{l c c c c}
    \toprule
    Method   &  $n_\text{Init}$ & $n_\text{Query}$  &  Penalized LogP ($\uparrow$ ) & Top 1 Penalized LogP ($\uparrow$ )\\
    \midrule
    MolDQN \citep{zhou2019MolDQN} & 250,000 & $\geq 5,000$ &  N/A & 11.84\\
    \midrule
    \multirow{2}{*}{W-LBO \citep{tripp2020}}   
        & 200 & 500 & 12.09±7.576  &  21.74\\ 
        & 250,000  & 500 & N/A & 27.84 \\
    \midrule
    \multirow{2}{*}{T-LBO \citep{grosnit2021tlbo}}   
       & 200  & 500 & 10.82±4.688  &  16.45\\ 
       & 250,000  & 500 & 26.11 & 29.06 \\
    \midrule
    \multirow{1}{*}{LOL-BO \citep{lol-bo}}
       & 250,000 & 500 & 27.53±2.393 & N/A \\
    \midrule
    \multirow{1}{*}{PG-LBO \citep{pg-lbo}}
       & 1000 & 500 & 23.23±2.913 & N/A \\
    \midrule
    \multirow{1}{*}{TSBO}  
      & 200 & 500 &\textbf{28.04±3.731} & \textbf{32.92} \\ 
    \bottomrule
    \end{tabular}
    \label{tab:500budget}
\end{table*}

\begin{table}[tb]
    \centering
    \caption{A broader comparison on the Expression Task}
    
    \begin{tabular}{l c c c}
    \toprule
    Method   &  $n_\text{Init}$ & $n_\text{Query}$  &  Expression ( $\downarrow$ )\\
    \midrule
    \multirow{2}{*}{W-LBO} 
        & 100 & 500 & 0.386±0.016 \\
        & 40000 & 500 & 0.314±0.1436 \\
    \midrule
    \multirow{1}{*}{T-LBO} 
        & 100 & 500 & 0.475±0.172 \\
    \midrule
    \multirow{1}{*}{PG-LBO} 
        & 100 & 500 & 0.358±0.195 \\
    \midrule
    \multirow{1}{*}{TSBO}  
      & 100 & 250 &\textbf{0.240±0.168} \\ 
    \bottomrule
    \end{tabular}
    \label{tab:expr-broader-compareison}
\end{table}



We employ multiple challenging blackbox optimization datasets to demonstrate  $\ourmethod$'s superior sample efficiency compared to recent baselines. Our results highlight that the proposed selective regularization forms the foundation of $\ourmethod$'s enhanced performance. Through detailed ablation studies, we quantify the distinct contributions of each component within $\ourmethod$,  providing insights into how these components collectively influence its overall performance.

\subsection{Experimental Settings} \label{sec:exp-setting}
We conduct experiments on three challenging high-dimensional global optimization tasks based on two datasets. The first dataset comprises 40,000 single-variable arithmetic expressions, and is employed for an \textbf{arithmetic expression reconstruction} task \citep{kusner2017grammar-vae}. The second  ZINC250K dataset \citep{sterling2015zinc}, consisting of 250,000 molecules, is used for two \textbf{chemical design} tasks with two objective molecule profiles: the penalized water-octanol partition coefficient (Penalized LogP) \citep{gomez2018semibo1} and the Ranolazine MultiProperty Objective (Ranolazine MPO) \citep{brown2019guacamol}. Detailed details of these tasks can be found in \cref{sec:appendix-dataset}.

\textbf{Baseline Methods}  $\ourmethod$ is benchmarked against 5 VAE-based latent space optimization baselines: LS-BO \citep{gomez2018semibo1}, W-LBO \citep{tripp2020}, T-LBO \citep{grosnit2021tlbo}, LOL-BO \citep{lol-bo}, and PG-LBO \citep{pg-lbo}. LS-BO performs BO in the latent space with a fixed pre-trained Variational Autoencoder (VAE) \citep{kingma2013vae}; W-LBO periodically fine-tunes the VAE with current labeled data; T-LBO introduces deep metric learning to W-LBO by additionally minimizing the triplet loss of the labeled data; LOL-BO optimizes the GP surrogate and VAE simultaneously via maximizing the joint likelihood; PG-LBO calibrates the VAE to improve the GP surrogate's accuracy on labeled data and to minimize the reconstruction loss on synthetic unlabeled data. Additionally, we include a random search algorithm Sobol \citep{owen2003sobol} and a reinforcement learning approach MOLDQN \citep{zhou2019MolDQN} for reference. 

\textbf{\ourmethod's Configurations} Two variants of \ourmethod\ are built on top of the baseline T-LBO \citep{grosnit2021tlbo}.
We denote \ourmethod\ with the optimized Gaussian distribution based unlabeled data sampling by \textbf{TSBO-Gaussian}, and that with the GEV distribution for sampling unlabeled data by \textbf{TSBO-GEV}. The full model setup can be found in \cref{sec:detail}. 



\subsection{Sample Efficiency of \ourmethod}\label{sec:result-main}


We set the initial amount of labeled data for starting off each BO run to 100 for the arithmetic expression reconstruction task and 200 for the two chemical design tasks, respectively.  \cref{fig:main_results} and \cref{tab:final} compare \ourmethod\ with four BO baselines and Sobol.  Both TSBO-GEV and TSBO-Gaussian consistently outperform T-LBO and other baselines across all tasks within  250 additional function evaluations (new data queries). Notably, TSBO-Gaussian drastically surpasses all baselines within the first 50 new queries, and finds further improved target values subsequently. These results underscore the superior sample efficiency of \ourmethod\ and the effectiveness of the proposed selective regularization.

We present a broader comparison on two optimization tasks with different data query budget settings.  \cref{tab:500budget} and \cref{tab:expr-broader-compareison} respectively show the results of the expression and chemical design task, where $n_\text{Init}$ is the number of initial labeled data used, and $n_\text{Query}$ the number of new queries.  \ourmethod\ attains the SOTA performance,
surpassing all baseline models and achieving a remarkable reduction in the utilization of labeled data by up to 364.2x.


\subsection{Generalization Improvement of Data Query GP Model of \ourmethod}

The generalization of the GP data query model is key to the overall BO performance. Here, we demonstrate improved generalization resulting from incorporating the pseudo labels predicted by the \ourmethod\ teacher as additional GP training data.
We assess the GP's accuracy across the whole search space (global) and in the high target value region (local) with a total of 100 test points. For the global assessment, we sample test data in the latent space from the VAE prior $\mathcal{N}(\bm{0},\bm{I})$. The local assessment samples test data from a Gaussian distribution centered at the point corresponding to the best target value found with a small standard deviation of $0.01$. The assessments are conducted after \ourmethod\ completes the final (250th) data query on the two Chemical Design tasks with 200 initial labeled molecules. 

\cref{tab:generalization} reports the negative log-likelihood (NLL) loss of the posterior prediction of a vanilla GP fitted exclusively on labeled data vs. that of the \ourmethod\ GP fitted on both the labeled data and unlabeled data with pseudo labels predicted by the teacher, both evaluated on the test data.  We use the abbreviations PL for pseudo-label, P-LogP for Penalized LogP, and R-MPO for Ranolazine MPO, respectively in the table. The \ourmethod\ GP  shows superior global and local generalization, both of which may boost \ourmethod's sample efficiency. Local GP generalization is more critical for Bayesian optimization as it helps query data with a higher target value. The improvement in local GP generalization brought by \ourmethod\ is more pronounced, specifically on the Penalized LogP task.


\subsection{Ablation Study} \label{sec:ablation}

We conduct an ablation study to assess the efficacy of various ingredients of $\ourmethod$, namely: 1) selective regularization to the teacher, 2) uncertainty awareness of the teacher-student model, 3) optimized unlabeled data sampling.
The experiments are conducted on the Chemical Design task, following the settings at the beginning of \cref{sec:result-main}. 

The results of the ablation study are presented in \cref{tab:full-ablation}, where we denote uncertainty awareness by ``UA".
The effectiveness of each of the three techniques listed above is manifested by a drastic performance drop resulting from its removal from \ourmethod.

In addition, \cref{sec:appendix-lambda} demonstrates the robustness of our approach to the value of the feedback weight $\lambda$, and \cref{sec:appendix-experiment-validation} shows the impact of the validation data selection.

\begin{table}[!tbp]
    \centering
    \caption{The NLL loss of data query model of \ourmethod\ on the Chemical Design Task}
    
    \begin{tabular}{l c c c}
    \toprule
    Data Query Model   & Test Region & P-LogP  & R-MPO \\
    \midrule
    GP w/o PL  & Global	& 0.881 & -1.504\\
    GP with PL & Global	& \textbf{0.863} & \textbf{-2.019}\\
    \midrule
    GP w/o PL  & Local	& 4.228 & -2.086\\
    GP with PL & Local	& \textbf{1.388} & \textbf{-2.391}\\
    \bottomrule
    \end{tabular}
    \label{tab:generalization}
\end{table}

\section{Related Work}



\paragraph{Semi-Supervised Learning (SSL)}
Beyond the scope of Bayesian optimization, many SSL techniques have been developed to serve the general goal of reducing expensive labeled data use \citep{ssl-flexmatch, ssl-usb}. SSL often involves consistency regularization \citep{ssl-PI-model}, maintaining the consistency of the model's predictions on unlabeled data under perturbations of either data \citep{ssl-vat, ssl-uda, ssl-mixmatch, ssl-remixmatch, ssl-fixmatch, ssl-softmatch} or the model \citep{ssl-temporal-ensembling, ssl-mean-teacher, ssl-noisy-student, pham2021meta_pseudo_label}. 
The family of model perturbation approaches makes use of two separate models with one acting as the teacher and the other a student, where the student learns from pseudo labels \citep{lee2013pseudo-labeling} predicted by the teacher \citep{pham2021meta_pseudo_label}. 
The proposed \ourmethod\ is the first work integrating one of the above SSL approaches, i.e., the teacher-student paradigm into BO while introducing the \emph{selective regularization} that is optimized for black-box global optimization. 

\paragraph{High-dimensional Bayesian Optimization in a Latent Space} 
Operating in a reduced dimensional latent space  \citep{latent-bo-combinatorial, latent-bo-decoder-uq} is essential for making BO applicable to high-dimensional optimization problems \citep{turbo-num1, high-bo-framework, high-bo-linear, high-bo-axis-align, high-bo-adaptive}. For this, it is a common practice to employ a dimension reduction model in the form of linear or nonlinear projections or VAE. 

Early latent space BO methods use a fixed dimension reduction model during the adaptive data query process. These models are either randomly initialized \citep{wang2016rembo}, or pre-trained on unlabeled data \citep{kusner2017grammar-vae, jin2018jtvae, alperstein2019all-smiles-vae}. 
Recent advances adapt the latent space, for example, by retraining the dimension reduction model with an additional loss defined on available labeled data alone \citep{eissman2018target-prediction, tripp2020, grosnit2021tlbo, lol-bo}, or in conjunction with sampled synthetic data \citep{chen2020silbo, pg-lbo}. W-LBO \citep{tripp2020} retrains the VAE-based dimension reduction with a weighted evidence lower bound (ELBO) on labeled data whose weight is proportional to the label value. T-LBO \citep{grosnit2021tlbo} introduces a triplet loss \citep{xing2002deep-metric-learning} to pull labeled data with similar target values together in the latent space. LOL-BO \citep{lol-bo} simultaneously optimizes the GP surrogate and VAE, 
and adopts the local approach TurBO \citep{turbo-num1} in the latent space. PG-LBO \citep{pg-lbo} updates the VAE using an MSE loss of the GP surrogate on labeled data cast in the latent space and weighted retraining \cite{tripp2020} on heuristically sampled unlabeled data with pseudo labels \citep{lee2013pseudo-labeling}. 
\begin{table}[tp]
    \centering
    \caption{\ourmethod's ablation on the Chemical Design Task}
    
    \begin{tabular}{l c c}
    \toprule
    TSBO Variant   &  Penalized LogP ($\uparrow$ )\\
    \midrule
    TSBO  w Random Sampler 
         &4.881±1.416\\
    TSBO  w/o UA   & 12.568±7.965\\    
    TSBO w/o Student's Feedback
         & 17.557±6.998\\

    TSBO  & 25.020±4.794 \\
    \bottomrule
    \end{tabular}
    \label{tab:full-ablation}
\end{table}

While the aforementioned approaches focus on optimizing dimension reduction, \ourmethod\ takes on an orthogonal approach to boost the generalization of the GP surrogate by utilizing high-quality ``pseudo" training data predicted by the well-regularized teacher model. The introduced selective regularization of the teacher model is instrumented by the uncertainty-aware teacher-student interaction and a systematically optimized unlabeled data sampler, all tailored for Bayesian optimization. It is worth noting that the dimension reduction techniques employed in the surveyed BO methods may be integrated into our \ourmethod\ framework.

\section{Discussion}\label{Discusion}
The proposed ($\texttt{TSBO}$) approach presents the first work integrating teacher-student based semi-supervised learning to enable sample-efficient Bayesian optimization. 
$\texttt{TSBO}$ incorporates uncertainty-aware teacher-student interactions and optimized unlabeled data sampling,  which collectively implement the selective regularization to the teacher. This makes it possible to enhance the generalization of the GP data query model by leveraging high-quality pseudo labels predicted by the teacher. \ourmethod\ achieves superior performance in comparison with other competitive latent space BO algorithms under tight labeled data budgets.

Opportunities exist for further improvement of \ourmethod\ in the future. For example, a more rigorous treatment of uncertainty quantification of the teacher-student model using techniques such as deep ensembles \citep{NIPS2017DeepEnsembles} and Bayesian neural networks \citep{Bayesian-Neural-Networks} may be explored to better mitigate the risk of predicted pseudo labels.


\bibliography{icml2024}
\bibliographystyle{icml2024}

\newpage
\appendix
\onecolumn



\section{Technical Details of Unlabeled Data Sampling}
\subsection{Optimized EVT-based Unlabeled Data Sampling}\label{sec:appendix-gev-sampling}

The GEV distribution \citep{fisher1928evt} is defined as
\begin{equation}
        p_{\y^*}(\y^*)=
        \mathbb{I}_{\{\xi\neq 0\}}\left(1+\xi \bar{\y}\right)^{-\frac{1}{\xi}} e^{-\left(1+\xi \bar{\y}\right) ^ {-\frac{1}{\xi}}} +
        \mathbb{I}_{\{\xi=0\}} e^{-\bar{\y}} e^{-e^{-\bar{\y}}},
\end{equation}
where $\bar{\y}:=\left(\y^*-a\right)/b$ defined by 3 learnable parameters of the GEV distribution: a location coefficient $a\in \R$,  a scale value $b>0$, and a distribution shape parameter $\xi\in\R$. 

We fit a GEV distribution $p_{\y^*}$ with parameters estimated by minimizing the NLL loss of several extreme labels. This GEV distribution captures the distribution of the best-observed target values as seen from the current evaluated data. As such, generating unlabeled data whose predicted labels follow the GEV distribution allows us to start out from the region of the existing extreme labeled data while exploring points with potentially even greater target values due to the random nature of the sampling process. Once the GEV distribution $p_{\y^*}$ is fitted, we adopt a specific MCMC method tailored for GEV \citep{hu2019gevmcmc} to sample from it.

\subsection{Unlabeled Data Sampling Distribution
Learned from Student’s Feedback} \label{sec:appendix-unlabel-update-rule}

We apply the reparametrization trick \citep{kingma2013vae} as our preferred sampling strategy. By introducing a random vector $\rv \in \mathcal{R} \subseteq \R^d$ and a mapping function $g(\cdot;\paramsU): \mathcal{R} \to \latentspace$, where $g(\rv;\paramsU) \sim \pdfZu$ when $\rv \sim \pdfrv$, we can efficiently sample unlabeled data $\zunlabel:= g(\rv; \paramsU)$ using $\pdfrv$, a known distribution that can be conveniently sampled from, such as a Gaussian distribution. 

Learning a parameterized sampling distribution by minimizing the feedback loss is a sensible choice. A large feedback loss is indicative of the use of unlabeled data with poor pseudo-label quality, which can potentially mislead the teacher-student model. We optimize $\paramsU$ to minimize the feedback loss $\lossfeedback$:
\begin{equation}
     \paramsU^* = \argmin_{\paramsU} \E_{\zsetunlabel \sim \pdfZu} \lossfeedback \big( \datasetlabel;  \paramsS, \zsetunlabel \big).
\end{equation}

The gradient for updating $\paramsU$ can be expressed using the reparametrization trick as follows:
\begin{equation}
    \nabla_{\paramsU} \E_{\zsetunlabel \sim \pdfZu} \lossfeedback \big( \datasetlabel;  \paramsS, \zsetunlabel \big) =\nabla_{\paramsU} \E_{\rvset \sim \pdfrv} \lossfeedback \big( \datasetlabel;  \paramsS , g(\rvset;\paramsU)\big)
\end{equation}
where $\rvset \in \R^{M\times d}$ is a batch of $M$ samples $\{\rv_{i}\}_{i=1}^{M}$. We incorporate the update of $\paramsU$ to the alternating one-step scheme for $\paramsS$ and $\paramsT$, as detailed in \cref{sec:appendix-rule}.

\section{Alternating One-step Update Rule} \label{sec:appendix-rule}
\begin{algorithm}[t]
\caption{Bi-Level Optimization of the Teacher-Student Model}\label{alg:ts-update}
\begin{algorithmic}
\STATE {\bfseries Input:}  
    Epochs $L$, feedback weight $\lambda$,
    teacher $\Teacher(\cdot; \paramsT^0)$, 
    student $\Student(\cdot; \paramsS^0)$,
    labeled data $\datasetlabel$, 
    unlabeled data $\zsetunlabel$
    
\STATE {\bfseries Output:}  Pseudo labels $\pseudolabels$

\FOR{$i = 1 \ \text{to}\  L$}
    \STATE Generate pseudo labels: $\pseudolabels \leftarrow \teacherPredMean(\zsetunlabel; \paramsT^{i-1})$ 
    \STATE Update the student model: $\paramsS^{i} \leftarrow \paramsS^{i-1} - \eta_{\Student}\cdot \nabla_{\paramsS^{i-1}} \lossunlabeled( \datasetunlabel(\paramsT^{i-1}) ;\paramsS^{i-1})$
    \STATE Fix $\paramsS^{i}$, and update the teacher model: $ \paramsT^{i} \leftarrow \paramsT^{i-1} - \eta_{\Teacher}\cdot \nabla_{\paramsT^{i-1}}\{\lambda\lossfeedback(\datasetlabel ;\paramsS^{i}) + \losslabeled (\datasetlabel ;\paramsT^{i-1})\}$
\ENDFOR
\STATE Predict pseudo labels:  $\pseudolabels \leftarrow \teacherPredMean(\zsetunlabel; \paramsT^{L})$
\end{algorithmic}
\end{algorithm}

To solve it in a computationally efficient way, we adopt an alternating one-step gradient-based approximation method from \citet{pham2021meta_pseudo_label}. When unlabeled data $\zsetunlabel$ are sampled from the distribution $\pdfZu (\cdot|\paramsU)$, we adopt the reparameterization trick to optimize $\paramsU$. In the $i\text{th}$ training iteration of the teacher-student, we update $\paramsU^{i-1}$ with a learning rate $\lrU$:

$\smallbullet\quad $Sample $\zsetunlabel$ with reparameterization trick: $\zsetunlabel = g(\rvset , \paramsU^{i-1})$ where $\rvset \sim \pdfrv$;

$\smallbullet\quad $Update the student model: $\paramsS^{i} \leftarrow \paramsS^{i-1} - \eta_{\Student}\cdot \nabla_{\paramsS^{i-1}} \lossunlabeled( \datasetunlabel(\paramsT) ;\paramsS^{i-1})$;
    
$\smallbullet\quad $Fix $\paramsS^{i}$, and update the teacher model: $ \paramsT^{i} \leftarrow \paramsT^{i-1} - \eta_{\Teacher}\cdot \nabla_{\paramsT^{i-1}}\{\lambda\lossfeedback(\datasetlabel ;\paramsS^{i}) + \losslabeled (\datasetlabel ;\paramsT^{i-1})\}$.

$\smallbullet\quad $Fix $\paramsS^{i}$, and update $\paramsU^{i}$: $\paramsU^{i} 
 \leftarrow \paramsU^{i-1} - \lrU  \nabla_{\paramsU^{i-1}}  \lossfeedback(\datasetlabel; \paramsS^{i})$.

\section{Dynamic Selection of Validation Data}\label{sec:appendix-validation}
It is attempting to use the full set of available labeled data $\datasetlabel$ as the validation set $\datasetval$ to assess the student, as proposed in \citep{pham2021meta_pseudo_label} for image classification proposes. However, it is not always optimal under the setting of BO, whose objective is to find the global optimum using an overall small amount of labeled data. Hence, the assessment of the student, which provides feedback to the teacher, shall be performed in a way to improve the accuracy of the teacher-student model in predicting the global optimum. Since the majority of labeled data $\datasetlabel$ are used in training the teacher, the quality of pseudo labels around $\datasetlabel$ is high.  Thus, validating the student using $\datasetlabel$ may lead to a low averaged loss, however, which is not necessarily indicative of the model's capability in predicting the global optimum. Our empirical study shows that the performance of $\ourmethod$ improves as the validation data are chosen to be the ones with higher target values. 
This is meaningful in the sense that assessing the student in regions with target values closer to the global optimum provides the best feedback to the teacher for improving its accuracy at places where it is most needed. 
We adopt a practical way to dynamically choose $\datasetval$ at each BO iteration: the subset of $\datasetlabel$ with the $\topk$ highest label values. For this, we apply a fast sorting algorithm to rank all labeled data. \cref{sec:appendix-experiment-validation} demonstrates the effectiveness of the proposed selection approach for $\datasetval$.

\section{Experimental Details}\label{sec:detail}


\subsection{High-dimensional Optimization Tasks}\label{sec:appendix-dataset}
\textbf{Arithmetic Expression Reconstruction Task:} The objective is to discover a single-variable arithmetic expression $\xopt = \texttt{1 / 3 + v + sin( v * v )}$. For an input expression $\x$, the objective function is a distance metric $f(\x)= \max\{-7, - \log(1+\text{MSE}(\x(\bm{\mathrm{v}})-\x^*(\bm{\mathrm{v}})) \}$, where $\bm{\mathrm{v}}$ are 1,000 evenly spaced numbers in $[-10, 10]$. A grammar VAE \citep{kusner2017grammar-vae} with a latent space of dimension 25 is adopted. It is pre-trained on a dataset of 40,000 expressions \citep{kusner2017grammar-vae}.

\textbf{Chemical Design Task:} The purpose of this task is to design a molecule with a required molecular property profile. The objective profiles considered are 1) the penalized water-octanol partition coefficient (Penalized LogP) \citep{gomez2018semibo1}, and 2) the Ranolazine Multiproperty Objective (Ranolazine MPO) \citep{brown2019guacamol}. A Junction-Tree VAE \citep{jin2018jtvae} with a latent space of dimension 56 and pre-trained on the ZINC250k dataset \citep{sterling2015zinc}.

For each task, prior to optimization, a VAE is pre-trained using unlabeled data through the maximization of the ELBO \citep{kingma2013vae}, and all methods employ this pre-trained VAE at the outset of optimization.

\subsection{\ourmethod's Model Architecture and Hyperparameters} \label{sec:appendix-hyperparam}
In \ourmethod, the teacher model is a multilayer perceptron (MLP) with 5 hidden layers and ReLU activation \citep{nair2010relu}. The output dimension is 2. The student model is a standard GP with an RBF kernel.

For the purpose of reproducibility, we provide a comprehensive account of the hyperparameters employed in all our experiments using \ourmethod. Our approach is based on T-LBO, and thus we adopt the default hyperparameters of VAE as suggested by \citep{grosnit2021tlbo}. The remaining hyperparameters, specific to \ourmethod, are presented in \cref{tab:hypers}. 

\begin{table}[!ht]
    \centering
    \begin{tabular}{l c c c }
    \hline
        Group & Hyper-parameter  & Expression & Chemical Design \\
    \hline
        \multirow{7}{*}{Common} & \makecell{Number of training steps in each BO iteration}  & 20 & 20 \\
        & Number of warm-up steps  & 2,000 & 2,000 \\
        & Feedback weight & $10^{-1}$ & $10^{-1}$ \\
        & Number of validation data &  10 & 30 \\
        & \makecell{Number of sampled unlabeled data} &  100 & 300 \\
        & Acquisition Function  & EI & EI \\
        & \makecell{Acquisition optimizer} &  LBFGS & LBFGS \\
    \hline
        \multirow{3}{*}{Teacher} & Learning rate  & $10^{-3}$ & $10^{-4}$ \\
        & \makecell{Batch size of labeled data} & 256 & 32 \\
        & Optimizer  & Adam & Adam \\
    \hline
        \multirow{4}{*}{Student} & Kernel  & RBF & RBF \\
        & Prior mean  & Constant & Constant \\
        & Learning rate  & $10^{-2}$ & $10^{-2}$ \\
        & Optimizer & Adam & Adam \\
    \hline
        \multirow{2}{*}{Data Query GP} & Kernel  & RBF & RBF \\
        & Prior mean  & Constant & Constant \\
    \hline
    \end{tabular}
    \caption{Hyper-parameters}
    \label{tab:hypers}
\end{table}


\subsection{Training of VAE in \ourmethod} \label{sec:appendix-tlbo-vae}
Although $\texttt{TSBO}$ stands as a general BO framework, it has been seamlessly integrated into T-LBO \citep{grosnit2021tlbo}, a state-of-the-art VAE-based BO method, to facilitate a fair comparison. The training approach for the VAE remains unaltered, aligning with T-LBO's methodology:

$\smallbullet\ $ Pretrain: Before the first BO iteration, the VAE is trained on the dataset in an unsupervised way to maximize the ELBO \citep{kingma2013vae};

$\smallbullet\ $ Fine-tune: After each $50$ BO iteration, the VAE is trained on all labeled data for 1 epoch to both maximize the ELBO and minimize the triplet loss which penalizes data having similar labels located far away. The weight of triplet loss is set to 10 in the Expression task and 1 in the Chemical Design task. 

The training schemes of all models proposed in $\texttt{TSBO}$ and the VAE are decoupled, rendering T-LBO an apt baseline for validating $\texttt{TSBO}$'s sample efficacy.

\begin{table}[!tbp]
    \centering
    \caption{The ablation test of the weight of the feedback loss }
    
    \begin{tabular}{l l c c c}
    \hline
    Method    & $\lambda$    &Expression ( $\downarrow$ ) & Penalized LogP ( $\uparrow$ ) & Ranolazine MPO ( $\uparrow$ )\\
    \hline
    T-LBO & - & 0.572±0.268 &5.695±1.254 & 0.620±0.043\\
    \hline
    \multirow{4}{*}{TSBO-Gaussian} 
    &0.001    & 0.240±0.168 & 21.106±8.960 & 0.713±0.021 \\
    &0.01    & 0.433±0.260 & 21.384±1.533 & 0.720±0.040\\
    &0.1   & 0.450±0.130 & 25.021±4.794  & 0.744±0.030\\
    
    & 1  & 0.474±0.113 & 21.122±7.494 & 0.712±0.023\\
    
    \hline
    \end{tabular}
    \label{tab:lambda}
\end{table}

\section{The Influence of the Feedback Weight in \ourmethod} \label{sec:appendix-lambda}
We analyze the influence of the selection of the feedback weight $\lambda$. Our experiments demonstrate that in a large range of $\lambda$, \ourmethod\ consistently outperforms the baseline T-LBO, underscoring the robustness of our approach to this hyper-parameter.

As shown in \cref{tab:lambda}, while the selection of $\lambda$ in $\{0.001. 0.01, 0.1, 1\}$ has an impact on \ourmethod's performance,  for each considered $\lambda$, TSBO-Gaussian consistently outperforms T-LBO, indicating that our success is not contributed to a deliberate $\lambda$ selection.

\section{Additional Ablation Test of Validation Data Selection} \label{sec:appendix-experiment-validation}
In order to verify the effectiveness of the proposed dynamic selection of validation data $\datasetval$ in \ourmethod, where $\datasetval$ is the subset of $\datasetlabel$ with the $K$ highest label values, we conduct an ablation study to compare it (TSBO-Gaussian) with two non-optimized $\datasetval$ selection strategies: random $K$ examples uniformly sampled from $\datasetlabel$ (TSBO-Gaussian-ValRand), and the current labeled dataset $\datasetlabel$ (TSBO-Gaussian-ValAll). These three variants of \ourmethod\ are measured on the Chemical Design task, and we report their average best values of 5 runs starting with 200 initial labeled data and a new data query budget of 250. $K$ is set to 30.

As shown in \cref{tab:val}, despite each variant of \ourmethod\ outperforms the baseline T-LBO, both TSBO-Gaussian-ValRand and TSBO-Gaussian-ValAll are less competitive than TSBO-Gaussian in finding the global maximum. This result meets our expectations: the student's feedback on those examples with the $K$ highest label values is more beneficial for the training of the teacher-student model, and eventually better facilitating the search for the maximum.

\begin{table}[!h]
    \centering
    \caption{Comparison of validation data selection of \ourmethod\ on the Chemical Design task}
    \begin{tabular}{l c c}
    \hline
        Method & Validation Data Selection & Penalized LogP ($\uparrow$ ) \\
    \hline
         T-LBO \citep{grosnit2021tlbo} & - & 5.70\\
    \hline
         TSBO-Gaussian-ValRand & Random 30 & 21.60 \\
         TSBO-Gaussian-ValAll & All labeled data & 22.65\\
         TSBO-Gaussian & Top 30 & \textbf{25.02}\\
    \hline
    \end{tabular}
    \label{tab:val}
\end{table}

\section{Robustness of \ourmethod\ to Noisy Labels}
In this section, we demonstrate the robustness of \ourmethod\ in noisy environments. We compare \ourmethod\ with the other baselines on the Chemical Design task, where all labels are subject to additional i.i.d. zero-mean Gaussian noises, with a standard deviation (std) of 0.1. All methods start with 200 initial labeled data and query 250 new examples. We report the mean and the standard deviation of the best values found by each method among 5 runs in \cref{tab:noisy}.
\begin{table}[!ht]
    \centering
    \caption{Labels with white Gaussian noises on the Chemical Design Task}
    \begin{tabular}{l c c}
    \hline
         \multirow{2}{*}{method} & \multicolumn{2}{c}{Best Penalized LogP ($\uparrow$ )}  \\
           & Noise variance=0 & Noise variance=0.1  \\ 
     \hline
         Sobol              &3.019±0.296 & 3.06±0.38 \\ 
         LS-BO              &4.019±0.366 & 4.41±0.68 \\ 
         W-LBO              &7.306±3.551 & 4.44±0.27 \\ 
         T-LBO              &5.695±1.254 & 4.22±0.68 \\ 
    \hline
        TSBO-GEV            &18.40±7.89 & 20.73±6.24 \\ 
        TSBO-Gaussian       &\textbf{25.02±4.794} & \textbf{25.97±4.82} \\ 
    \hline
    \end{tabular}
    \label{tab:noisy}
\end{table}

While nearly all of the baselines, especially T-LBO, exhibit decreases in the noisy scenario compared to the results obtained without observation noise, \ourmethod\ shows no performance deterioration. This phenomenon demonstrates the noise-resistant capability of the proposed uncertainty-aware teacher-student model.

\section{Broader Impact}
The proposed \ourmethod\ has the potential for significant positive impacts in various domains. By effectively finding the optimum compared with baselines on multiple datasets, \ourmethod\ offers a promising solution to enhance the efficiency and efficacy of optimization processes given limited labeled data and evaluation budgets. For instance, in engineering and manufacturing, \ourmethod\ can facilitate outlier detection, failure analysis, and the design of more efficient processes, leading to increased productivity and reduced resource consumption. By enabling faster and more accurate optimization, \ourmethod\ can ultimately benefit society as a whole.

Even though \ourmethod\ holds great promise, it is important to acknowledge and mitigate potential negative impacts. One concern is the overreliance on automated optimization algorithms, which could lead to a decreased emphasis on human intuition and creativity. \ourmethod\ should be used as a supportive tool that enhances human decision-making rather than replacing it entirely. Additionally, there is a risk of bias in the optimization process if the training data used for the teacher model contain inherent biases. Careful attention must be given to the training data to ensure fair and unbiased optimization results.

In conclusion, \ourmethod\ offers significant potential for broad impact in optimization tasks. By improving the efficiency and efficacy of optimization processes, \ourmethod\ can accelerate the discovery of optimal solutions, benefiting various industries and ultimately improving the well-being of individuals and society at large. However, it is important to consider and mitigate potential negative impacts, such as overreliance on automation and the risk of bias, to ensure that \ourmethod\ is used responsibly and ethically. With proper safeguards and considerations, \ourmethod\ can be a valuable tool that enhances human expertise and drives advancements in optimization across diverse domains.


\end{document}